\documentclass{article}

\usepackage{arxiv}

\usepackage[utf8]{inputenc} 
\usepackage[T1]{fontenc}    
\usepackage{hyperref}       
\usepackage{url}            
\usepackage{booktabs}       
\usepackage{amsfonts}       
\usepackage{nicefrac}       
\usepackage{microtype}      
\usepackage{lipsum}
\usepackage{cite} 
\usepackage{graphicx} 
\usepackage{float} 
\graphicspath{ {./images/} }
\usepackage[cmex10]{amsmath}
\title{A Generative Adversarial Network for AI-Aided Chair Design}

\author{
  Zhibo Liu\thanks{Zhibo Liu and Feng Gao are co-first authors.} \\
  National Engineering Lab for Video Technology\\
  Peking University,Beijing, China \\ 
  \texttt{zhibo-liu@outlook.com} \\
  \And
  Feng Gao \\
  The Future Lab, Tsinghua University, Beijing, China \\
  \texttt{gaofeng2018@tsinghua.edu.cn} \\
  \AND
  Yizhou Wang \\
  National Engineering Lab for Video Technology \\
  Peking University,Beijing, China \\ 
  \texttt{Yizhou.Wang@pku.edu.cn} \\
}

\begin{document}
\maketitle

\begin{abstract}
We present a method for improving human design of chairs. The goal of the method is generating  enormous chair candidates in order to facilitate human designer by creating sketches and 3d models accordingly based on the generated chair design. It consists of an image synthesis module, which learns the underlying distribution of training dataset, a super-resolution module, which improve quality of  generated image and human involvements. Finally, we manually pick one of the generated candidates to create a real life chair for illustration. 
\end{abstract}


\section{Introduction}
Leveraging the power of artificial intelligence to improve design has been becoming an active research field ~\cite{pix2pix2017,CycleGAN2017,liu2017unsupervised,huang2018munit}. Different design projects require different personal experience, theories and skills. In this paper, we specifically target our problem on that how to achieve better chair design in the age of deep neural network?

The first natural question is that how to define a good design of chair? We believe that a good chair design needs to balance the practicality and creativity. In other words, people can sit on it while enjoying its novel style which consists of both material (texture) and structure (shape). Therefore, the network must learn the underlying distribution of chairs so that the samples drawn from it can meet the practical requirement. 

The second question is that how to leverage the power of artificial intelligence, especially deep learning \cite{lecun2015deep} methods, to enhance human creativity and inspiration? Neural style transfer \cite{gatys2016image,johnson2016perceptual} separates image content from style explicitly so that the network is able to apply style from the style image to a target image. Image to image translation learns a projection from source domain to target domain \cite{CycleGAN2017,huang2018munit}, which can also be treated as learning style representation in the context of artistic design. However, separating style from content can only help designers when the shape (content) of the chair is given. Designing a good chair from scratch implies generating both content and style. Therefore, it's necessary to learn the underlying distribution of chairs.

In this paper, we propose a method for improving human design of chairs by generating enormous chair design candidates sampled from the learned underlying distribution of training data, which can speed up the designing process tremendously while maintaining the variety of shapes and textures. After that, human designers can select from chair design candidates and turn them into reality, as shown in Fig.1. The contribution of this paper is twofold:

\begin{enumerate}
\item We present a deep neural network for improving human design of chairs which consists of an image synthesis module and a super-resolution module. 

\item We select one of the candidates as design prototype and create a real life chair based on it. To the best of our knowledge, this is the first physical chair created with the help of deep neural network, which bridges the gap between AI and design. 
\end{enumerate}

\begin{figure*}[h]
    \includegraphics[width=\textwidth]{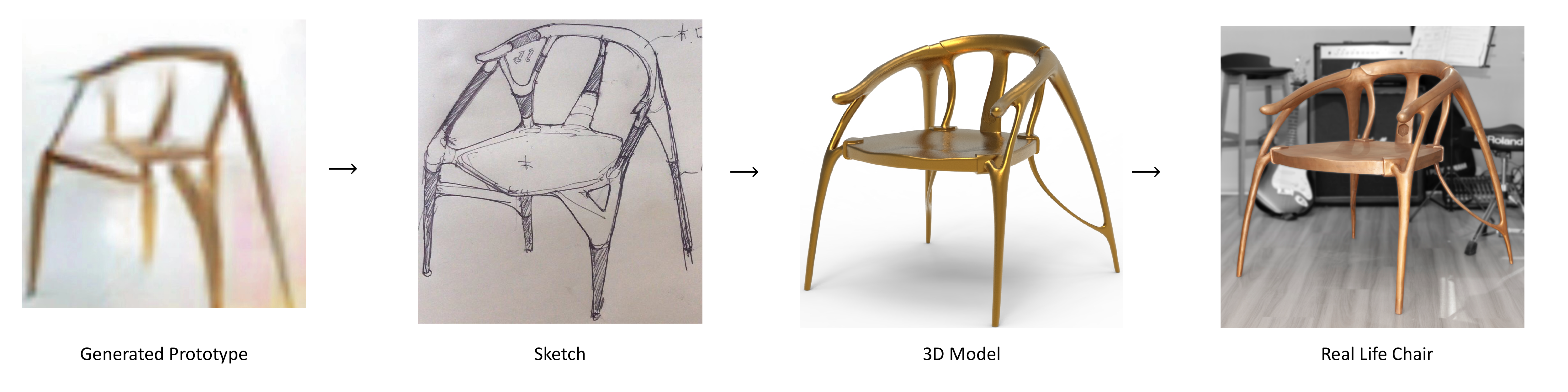}
    \caption{The pipeline of turning a chair design prototype selected from 320,000 generated image candidates into sketch, 3D model and finally a real life chair.}
\end{figure*}

\section{Related Works}
\subsection{Generative Adversarial Network}

Generative adversarial network \cite{goodfellow2014generative} has shown us impressive results on image generation \cite{radford2015unsupervised,karras2017progressive,brock2018large,karras2018style}. During training process, a generator tries to fool a discriminator which in turn aims to distinguish between generated images and real images. With adversarial loss \cite{goodfellow2014generative} that guides the training process, the generator is ultimately capable of learning the underlying distribution of training data implicitly. Various improvements have been made to improve GAN's performance by adopting better objective functions \cite{arjovsky2017wasserstein,gulrajani2017improved}, and novel network architectures \cite{karras2017progressive,brock2018large,karras2018style}. In this paper, we apply GAN to learn the chair dataset distribution and the high resolution prior.  
\subsection{Image to Image Translation}

Image to Image translation methods have been successfully applied for style transfer and image super-resolution, which can be used as a powerful tool for auxiliary design.For style transfer task, CNN based methods learn desired texture from style domain and transfer it to content domain \cite{gatys2016image,johnson2016perceptual}.GAN based methods learn projections between two domains by cycle consistency loss \cite{CycleGAN2017,kim2017learning,yi2017dualgan} or by sharing latent space features \cite{liu2017unsupervised,huang2018munit,DRIT,NIPS2018_7404,Lin2018ConditionalIT}. The application of those methods can be applied for improving human design. For example, a designer can draw a sketch of skirt and let the network to fill the texture. However, such methods don't meet our requirement for the reason that our goal is to design chairs with both creative shapes and textures.  Simply altering color and material of chairs can't help designers. Image super-resolution can be viewed as an image to image translation problem that the goal is to learn projection from low resolution domain to high resolution domain. Johnson \cite{johnson2016perceptual} introduce perceptual loss \cite{gatys2016image} to CNN based method. Meanwhile, recent GAN based method produce state-of-the-art performance \cite{Ledig2017PhotoRealisticSI,Wang2018ESRGANES}.

\section{Methods}
In order to generate chairs with sufficient shape and texture variation while maintaining high resolution, we introduce the image synthesis module and the super-resolution module to solve those problem respectively. Finally, human involvements are necessary to turn a chair design prototype and a sketch into reality.  The complete pipeline of our proposed method is illustrated in Figure 2. 

\begin{figure*}[h]
    \includegraphics[width=\textwidth]{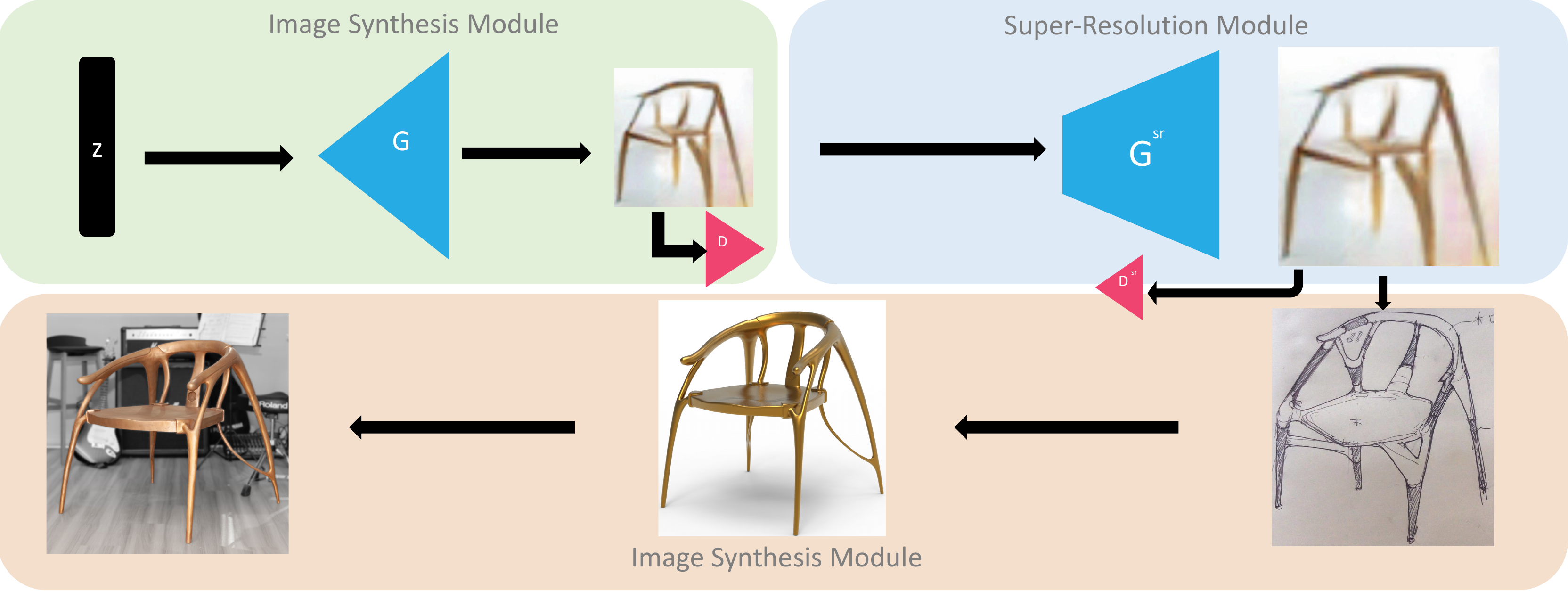}
    \caption{The network architecture of our proposed method which consists of the image synthesis module, super-resolution module and human involvement. Two modules are trained separately and the super-resolution module serves as a high resolution image prior which is fine-tuned on newly created chair dataset.}
\end{figure*}

\subsection{Image Synthesis Module}
The goal of image synthesis module is to learn chair data distribution and sample from it.  We denote chair data distribution as $x \sim p_{chair}(x)$ and 100-dimensional Uniformed random noise $z \sim p_Z(z)$. We apply adversarial loss \cite{goodfellow2014generative} to constrain the training process. For generator $G$ and discriminator $D$, we express the objective as: 
\begin{equation} \label{eq1}
\begin{split}
\mathcal{L}_{GAN}^{gen}(G,D) =\min_{G}\max_{D} E_{x\sim p_{chair}(x)}[\log D(x)] +\\ E_{z\sim p_{Z}(z)}[\log (1-D(G(z)))]
\end{split}
\end{equation}

Where $G$ tries to generate image $G(z)$ that is indistinguishable from discriminator $D$. At the same time, the discriminator $D$ aims to distinguish between real chair samples $x$ with generated images $G(x)$. Therefore, $G$ tries to minimize the objective while $D$ tries to maximize it. With the guidance of adversarial loss and the power of fully convolutional networks, the generator can learn the shape and texture of chair distribution ultimately. Figure 3 show the generated images.


\subsection{Super-Resolution Module}
The super-resolution module serves as a higher resolution image prior which learns the projection from low quality images to their corresponding high quality images. Since we specifically target our task on chair image generation, in order to obtain a better generated chair quality, we create a low and high resolution dataset of chairs by downscaling the corresponding  image. 

We apply adversarial loss and perceptual content loss \cite{gatys2016image,johnson2016perceptual} to generate higher resolution images. Let $y_{hr}$ and $y_{lr}$ be images from high resolution domain and corresponding low resolution domain. $G_{sr}$ and $D_{sr}$ be the generator and discriminator of super-resolution module. The adversarial loss function is: 
\begin{equation} \label{eq2}
\begin{split}
\mathcal{L}_{GAN}^{sr}(G_{sr},D_{sr}) =\qquad \qquad \qquad \qquad \qquad \qquad \qquad\\ \min_{G_{sr}}\max_{D_{sr}} E_{y_{hr}\sim p_{train}(y_{hr})}[\log D_{sr}(y_{sr})] +\\ E_{y_{lr}\sim p_{G_{sr}}(y_{lr})}[\log (1-D_{sr}(G_{sr}(y_{lr})))]
\end{split}
\end{equation}
The difference between a high resolution image $y_{hr}$ and super-resolution image $G_{sr}(y_{lr})$ is measured by perceptual content loss given a pretrained VGG19 network \cite{simonyan2014very}:
\begin{equation} \label{eq3}
\begin{split}
\mathcal{L}_{content/i,j}^{sr}(G_{sr}) =\frac{1}{W_{i,j}H_{i,j}}\sum_{k=1}^{W_{i,j}}\sum_{l=1}^{H_{i,j}}\big\{\phi_{i,j}(y_{hr})_{k,l}\\ - \phi_{i,j}[G_{sr}(y_{lr})]_{k,l}\big\}^2
\end{split}
\end{equation}
Where $\phi_{i,j}$ is the feature response of $i^{th}$ layer and $j^{th}$ filter. $W_{ij}$ and $H_{ji}$ indicate  the dimensions of the respective feature maps of VGG19 network. 

\begin{figure}[h]
    \centering
    \includegraphics[width=0.48\textwidth]{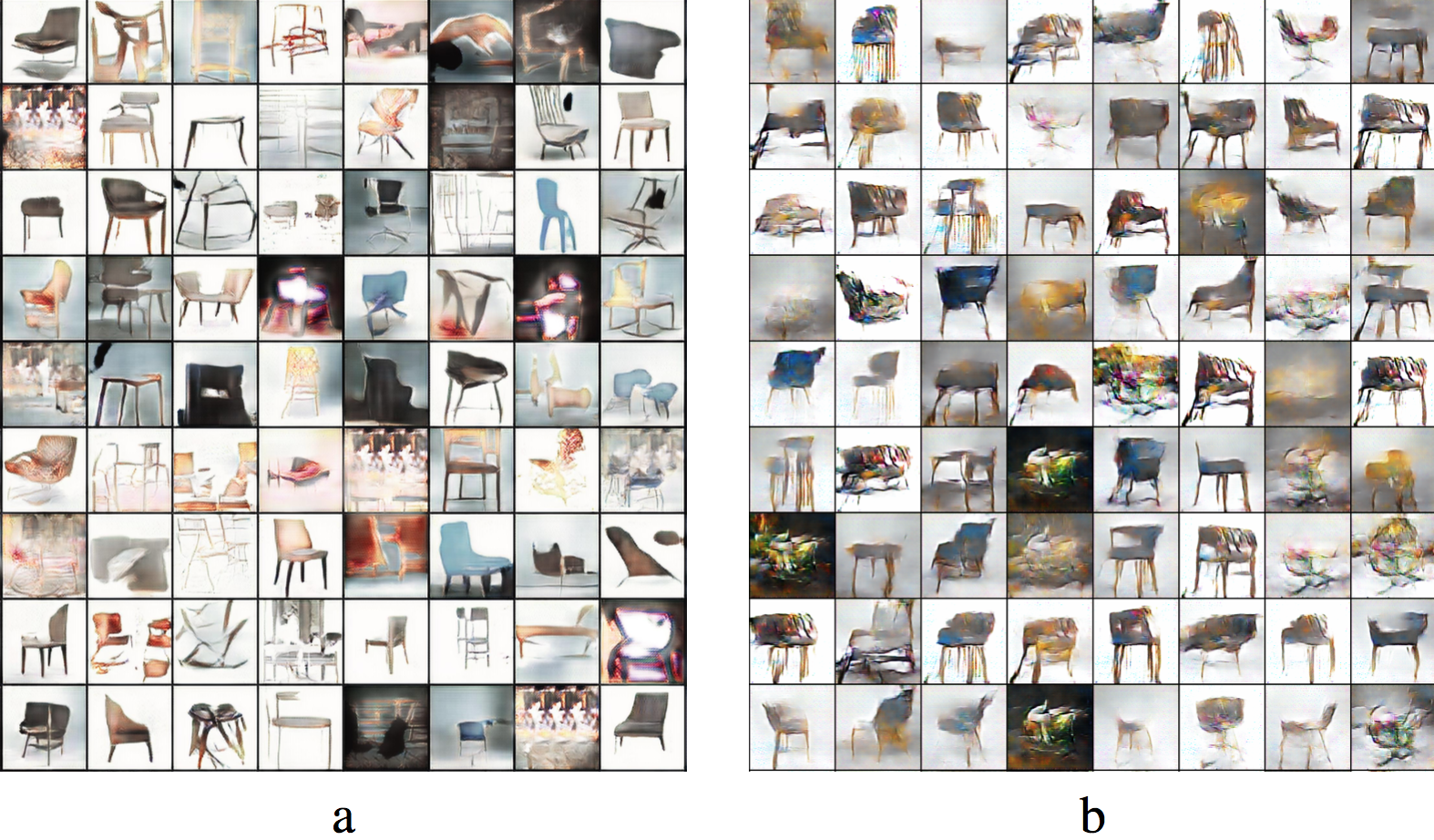}
    \caption{a) Generated chair design candidates at $256 \times 256$ resolution. The training dataset size is around 38,000  b) Generated chair images with 1000 training data. Increasing training dataset size leads to better learned implicit distribution which is represented by sampled images.}
\end{figure}

\subsection{AI Aided Chair Design and Realization }
It is necessary for human involvements that turn a generated 2d image prototype into a sketch, a 3d model and ultimately a real life chair. After the generation of 320,000 chair candidates, we spend few ours on final chair prototype selection. Compared with traditional time-consuming chair design process, our method speed up the process dramatically while keeping creativity and originality. 


\begin{figure}[h]
    \centering
    \includegraphics[width=0.45\textwidth]{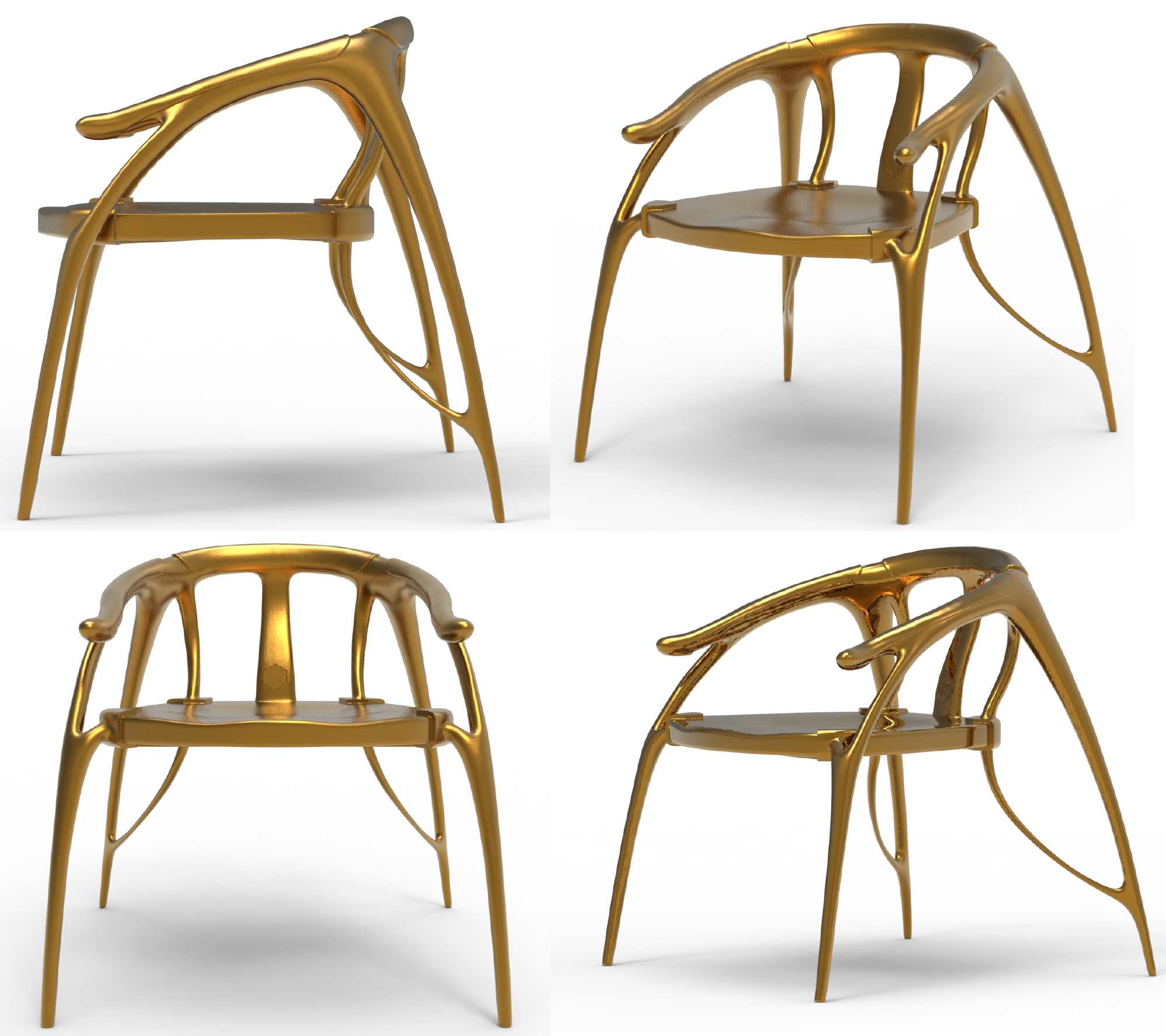}
    \caption{3D model created based on AI-aided chair design prototype.}
\end{figure}

\section{Implementations}
\textbf{Dataset:}  Around 38,000 chair images scraped from Pinterest for the image synthesis module. We downscale the images by a factor of 4 for fine-tuning the image super-resolution network.

\textbf{Network Architecture:}  The overall architecture is shown in Fig 2. For image synthesis we adopt DCGAN \cite{radford2015unsupervised} architecture whose generator consists of a series of 4 fractional-strided convolutions with batch normalization \cite{ioffe2015batch} and rectified activation \cite{nair2010rectified} applied. And the discriminator has a mirrored network structure with strided convolutions and leaky rectified activation \cite{maas2013rectifier} instead. For super-resolution we adopt the architecture of SRGAN \cite{Ledig2017PhotoRealisticSI} with  $i=5, j=4$ for perceptual content loss $\mathcal{L}_{content/5,4}^{sr}$ . 

\textbf{Training:}  We used 2 GTX1080Ti GPUs to train the image synthesis module for 200 epochs with mini-batch size of 128, learning rate of 0.0002 and Adam \cite{kingma2014adam} with $\beta_1 = 0.9$ . Two modules are trained separately. In order to improve super-resolution performance on chair data, we fine-tuned SRGAN on downscaled chair dataset.


\section{Results}
We generate 320,000 chair design candidate at $256 \times 256$ resolution (Fig 3) and spend few hours on selecting candidates as prototype. Finally, a real life chair is made based on the 2d sketch and 3d model of a prototype (Figure 5).

\begin{figure}[h]
    \centering
    \includegraphics[width=0.45\textwidth]{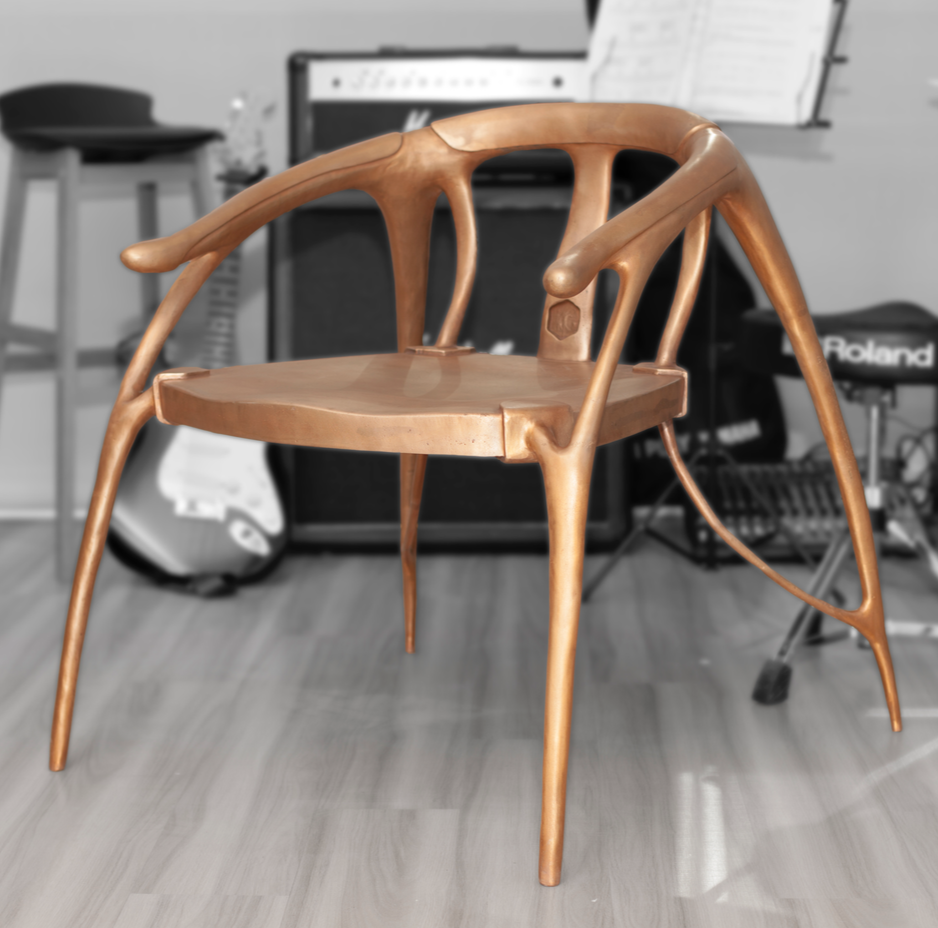}
    \caption{Real life chair created based on AI-aided chair design prototype.}
\end{figure}


\section{Conclusion}
We proposed a method for improving chair design, which consists of an image synthesis module, a super-resolution module and human involvement. It can speed up the designing process tremendously while balancing the practicality and creativity. The generated chair design candidates shows variety of shapes and textures. We finally illustrate our method by selecting one of the candidates as design prototype and create an AI-aided real life chair.

\bibliographystyle{unsrt}  
\bibliography{references}  

\end{document}